 \documentclass[sigconf]{acmart}

\AtBeginDocument{%
  \providecommand\BibTeX{{%
    \normalfont B\kern-0.5em{\scshape i\kern-0.25em b}\kern-0.8em\TeX}}}

\copyrightyear{2024}
\acmYear{2024}
\setcopyright{acmlicensed}
\acmConference[ICMR '24]{Proceedings of the 2024 International Conference on Multimedia Retrieval}{June 10--14, 2024}{Phuket, Thailand}
\acmBooktitle{Proceedings of the 2024 International Conference on Multimedia Retrieval (ICMR '24), June 10--14, 2024, Phuket, Thailand}
\acmDOI{10.1145/3652583.3658109}
\acmISBN{979-8-4007-0619-6/24/06}
\renewcommand\footnotetextcopyrightpermission[1]{} 
\usepackage{multirow}
\usepackage{enumitem}
\usepackage{balance}
\usepackage{fancyhdr}
\usepackage{pifont}
\begin{document}

\title{Enhancing Visible-Infrared Person Re-identification with Modality- and Instance-aware Visual Prompt Learning}


\author{Ruiqi Wu}
\orcid{0009-0003-5171-4548}
\authornote{ These authors contributed equally to this work.}
\affiliation{
 \institution{School of Computer Science, Northwestern Polytechnical University}
 \city{Xi'an}
 \state{Shaanxi}
 \country{China}
}
\affiliation{
 \institution{Ningbo Institute, Northwestern Polytechnical University}
 \city{Ningbo}
 \state{Zhejiang}
 \country{China}
}
\affiliation{
 \institution{National Engineering Laboratory for Integrated Aero-Space-Ground-Ocean Big Data Application Technology}
 \city{Xi'an}
 \state{Shaanxi}
 \country{China}
}
\email{wurq@mail.nwpu.edu.cn}

\author{Bingliang Jiao}
\authornotemark[1]
\orcid{0000-0001-5994-6181}
\affiliation{
 \institution{School of Computer Science, Northwestern Polytechnical University}
 \city{Xi'an}
 \state{Shaanxi}
 \country{China}
}
\affiliation{
 \institution{Ningbo Institute, Northwestern Polytechnical University}
 \city{Ningbo}
 \state{Zhejiang}
 \country{China}
}
\affiliation{
 \institution{National Engineering Laboratory for Integrated Aero-Space-Ground-Ocean Big Data Application Technology}
 \city{Xi'an}
 \state{Shaanxi}
 \country{China}
}
\email{bingliang.jiao@mail.nwpu.edu.cn}

\author{Wenxuan Wang}
\orcid{0000-0001-7142-9734}
\authornote{ Corresponding author.}
\affiliation{
 \institution{School of Computer Science, Northwestern Polytechnical University}
 \city{Xi'an}
 \state{Shaanxi}
 \country{China}
}
\affiliation{
 \institution{Ningbo Institute, Northwestern Polytechnical University}
 \city{Ningbo}
 \state{Zhejiang}
 \country{China}
}
\affiliation{
 \institution{National Engineering Laboratory for Integrated Aero-Space-Ground-Ocean Big Data Application Technology}
 \city{Xi'an}
 \state{Shaanxi}
 \country{China}
}
\email{wxwang@nwpu.edu.cn}



\author{Meng Liu}
\orcid{0009-0008-1202-3767}
\affiliation{
 \institution{Honors College, Northwestern Polytechnical University}
 \city{Xi'an}
 \state{Shaanxi}
 \country{China}
}
\affiliation{
 \institution{School of Electronics and Information, Northwestern Polytechnical University}
 \city{Xi'an}
 \state{Shaanxi}
 \country{China}
}
\email{2020300029@mail.nwpu.edu.cn}

\author{Peng Wang}
\orcid{0000-0002-9218-9132}
\affiliation{
 \institution{School of Computer Science, Northwestern Polytechnical University}
 \city{Xi'an}
 \state{Shaanxi}
 \country{China}
}
\affiliation{
 \institution{Ningbo Institute, Northwestern Polytechnical University}
 \city{Ningbo}
 \state{Zhejiang}
 \country{China}
}
\affiliation{
 \institution{National Engineering Laboratory for Integrated Aero-Space-Ground-Ocean Big Data Application Technology}
 \city{Xi'an}
 \state{Shaanxi}
 \country{China}
}
\email{peng.wang@nwpu.edu.cn}

\renewcommand{\shortauthors}{Ruiqi Wu, Bingliang Jiao, Wenxuan Wang, Meng Liu, \& Peng Wang}

\begin{abstract}
\renewcommand{\thefootnote}{**}
\footnotetext{This preprint follows the arXiv.org perpetual, non-exclusive license copyright agreement. The final version of this work has been accepted for publication in ICMR'24 by ACM, adhering to the  "ACM licensed" copyright agreement. When citing, please refer to the ACM Reference Format.}

The Visible-Infrared Person Re-identification (VI ReID) aims to match visible and infrared images of the same pedestrians across non-overlapped camera views. 
These two input modalities contain both invariant information, such as shape, and modality-specific details, such as color. An ideal model should utilize valuable information from both modalities during training for enhanced representational capability. 
However, the gap caused by modality-specific information poses substantial challenges for the VI ReID model to handle distinct modality inputs simultaneously. 
To address this, we introduce the Modality-aware and Instance-aware Visual Prompts (MIP) network in our work, designed to effectively utilize both invariant and specific information for identification. 
Specifically, our MIP model is built on the transformer architecture. In this model, we have designed a series of modality-specific prompts, which could enable our model to adapt to and make use of the specific information inherent in different modality inputs, thereby reducing the interference caused by the modality gap and achieving better identification. 
Besides, we also employ each pedestrian feature to construct a group of instance-specific prompts. These customized prompts are responsible for guiding our model to adapt to each pedestrian instance dynamically, thereby capturing identity-level discriminative clues for identification. 
Through extensive experiments on SYSU-MM01 and RegDB datasets, the effectiveness of both our designed modules is evaluated. Additionally, our proposed MIP performs better than most state-of-the-art methods.

\end{abstract}

\keywords{Cross-Modality Person Re-Identification, Visible-Infrared Person Re-Identification, Visual Prompt Learning}





\maketitle

\begin{figure}[htbp] 
	\centering
 \includegraphics[width=0.95\linewidth,scale=0.95]{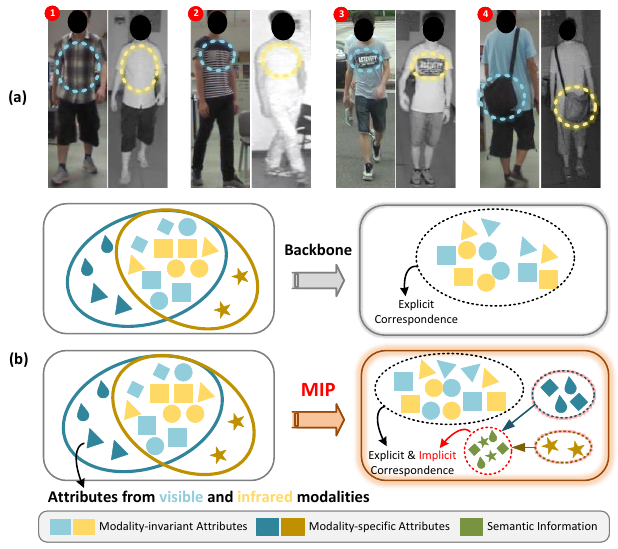}
        \vspace{-2mm}
	\caption{The illustration of our motivation. (a) Our motivation is to utilize modality-specific details to reveal potential relationships. For example, characteristics like clothing types/materials, discerned from visible images, could influence the heat radiation intensities, thereby impacting infrared brightness variations. 
    For intuitive example, in case 1, a uniform brightness is expected as the striped plaid shirt suggests similar materials.  Conversely, Case 3 exhibits varied brightness due to different materials in the T-shirt and logo. (b) Unlike traditional methods emphasizing modality-invariant information while overlooking modality-specific information, our approach integrates modality-specific attributes like color, texture, and brightness to extract and explore potential relationships.}
	\label{fig1}
\end{figure}

\section{Introduction}

Person Re-Identification (ReID) aims to retrieve images of the same individuals across different cameras. Driven by its applications in public security and video surveillance, ReID has garnered significant attention and achieved notable progress. 
Many existing methods~\cite{zheng2015scalable, qian2018pose, sun2018beyong, YEMANG2021AGWdeep} merely focus on the re-identification in daytime scenarios while ignoring the low-light conditions. However, treating ReID as a single-modality problem is unreasonable, which inevitably leads existing methods to fall short in low-light conditions.
To address this, infrared cameras have been introduced to capture images for all-day surveillance, leading to the emergence of the Visible-Infrared Person Re-Identification (VI ReID) task.

Different from the visible-visible ReID task, the query and gallery sets in VI ReID are captured by cameras with distinct modalities, resulting in significant modality gaps among the compared person images. 
To address these modality discrepancies, a straightforward approach involves modality transfer via Generative Adversarial Networks (GAN)~\cite{Goodfellow2014Generative}. Typically, GAN-based methods~\cite{dai2018cmgancross, wang2019alignganrgb, WANG2019D2RLLEARNING, liu2022tsmerevisiting} are designed to transform images from one modality to another, achieving effective cross-modality matching. However, challenges arise because the brightness of infrared images may not perfectly correspond to the color of visible images~\cite{LIANG2023CMTRTMM}.
A more general approach involves extracting discriminative modality-invariant information. For instance, SPOT~\cite{CHEN2022SPOTstructure} employs complex multi-level alignment mechanisms and leverages physics knowledge, such as human body structure, to learn discriminative cross-modality invariant features.

Although the existing methods could alleviate the interference caused by the modality gap, they focus on simply eliminating modality-specific information, overlooking the potential utilization of such information.
This is unreasonable as there could exist potential correspondences between modality-specific information that could also enhance re-identification~\cite{peng2018modality, WU2020modality}. 
For example, from the clothing types and material, which could be identified from visible images, we can briefly infer the heat radiation intensity of the clothing's surface. This factor directly influenced the brightness of the corresponding area in infrared images.
Intuitively, from the visible image in the first case of Figure~\ref{fig1}(a), the pedestrian wears a striped plaid shirt and the material of the different stripes appears to be roughly the same. Based on this, we could expect uniform heat radiation intensities, i.e., uniform brightness in the infrared image, of this upper dress.

Based on this insight, in this work, we aim to make use of both modality-invariant and modality-specific information to achieve better identification, as shown in Figure~\ref{fig1}(b).
This step is non-trivial, as the modality-specific information inevitably causes the semantic divergence between distinct modality inputs.
This divergence makes it challenging for the model to adapt to and process different modality inputs simultaneously. Meanwhile, using separate independent models for different modality inputs is also sub-optimal, as it struggles to fully utilize modality-invariant information, leading to insufficient representational capability.
Here, we notice that visual prompts could be a good tool to address this issue.
Its extensive use in numerous existing works~\cite{jia2022visual, YANG2022prompting, zhu2023visual, lu2022prompt, Bahng2022exploring, chen2022AdaptFormer} showcases its ability to preserve the foundational knowledge inherent in the backbone model while adapting models efficiently to various tasks.
This inspires us to leverage the visual prompts to learn the modality-invariant information with the backbone model and use a group modality-specific prompt to help the model flexibly adapt and make use of the modality-specific information. 
Besides, in this work, we also employ each pedestrian feature to construct a group of instance-specific prompts. These customized prompts are responsible for guiding our model to adapt to each pedestrian instance dynamically, thereby capturing discriminative clues for identification. 

Specifically, in this paper, 
we propose a novel and effective Modality-aware and Instance-aware Visual Prompts (MIP) network. The MIP network comprises a global backbone and two prompt learning modules: a Modality-aware Prompt Learning (MPL) module and an Instance-aware Prompt Generator (IPG) module. 
Initially, we design the MPL module with several learnable prompt vectors. Depending on the modality of inputs, the MPL module provides the backbone with the corresponding modality-specific prompt. Here, the backbone is trained to extract cross-modality consistency information, while the modality-specific prompt guides the backbone to adapt to inputs of different modalities and utilize modality-specific information.
Moreover, we have devised an innovative IPG module based on transformer architecture. In this module, we employ a transformer layer to transfer the identity-specific information from the preliminarily extracted pedestrian features into a group of learnable vectors to construct instance-specific prompts. These prompts are then inserted into the backbone model to direct it to dynamically adapt to the current instance, thereby capturing discriminative clues for identification.

We summarize the contribution of our work as follows:
\setlist[itemize]{topsep=1pt, partopsep=1pt, parsep=0pt, itemsep=1pt}
\begin{itemize}
  \item We propose a novel VI ReID method, the Modality-aware and Instance-aware Visual Prompts (MIP) network, incorporating visual prompt learning into the VI ReID field.
  \item We design a Modality-aware Prompt Learning (MPL) module and an Instance-aware Prompt Generator (IPG) to generate modality-specific and instance-specific prompts for the ReID model, which guide the model to adapt to inputs of different modalities and leverage modality-specific information, and capture identity-specific discriminative clues that help re-identification, respectively.
  \item We execute extensive experiments on VI ReID benchmarks SYSU-MM01 and RegDB, which validate the effectiveness of both our designed modules and demonstrate that MIP performs better than most state-of-the-art methods.
\end{itemize}

\section{Related Work}
In this section, we briefly review the existing algorithms related to our work in the areas of person re-identification, visible-infrared person re-identification, and visual prompt learning. 
\subsection{Person Re-Identification}
Person Re-identification (ReID) is a pivotal task that involves matching query person images with target person images from a gallery image set, attracting considerable attention for its real-world applications and prompting the development of diverse methodologies~\cite{zheng2015scalable,wu20193d,luo2020atrong,sun2018beyong,wang2018learning,dong2014deepmetric,qian2020long,Hermans2017indensetriplet, chen2017beyongdquaruplet, jiao2022dynamically, jiao2022Generalizable}. Similar tasks include vehicle ReID~\cite{WANG2019Vehicle, liu2016large}, animal ReID~\cite{jiao2023towards}, etc.

ReID methods typically comprise two key components: feature representation learning and deep metric learning. 
Global-based approaches like VLAD~\cite{wu20193d}, BNNeck~\cite{luo2020atrong}, among others, have been introduced to extract global-level feature representations for individuals' images. 
Furthermore, part-based methods, such as PCB-RPP~\cite{sun2018beyong}, leverage part-level clues to amalgamate more robust representations for retrieval purposes. 
Some algorithms combine both global and local features to exploit their respective advantages. For instance, Wang et al.~\cite{wang2018learning} proposed a multiple granularity network with one branch for global feature representation and two branches for local feature representation. 
Deep metric learning techniques~\cite{dong2014deepmetric, Hermans2017indensetriplet, chen2017beyongdquaruplet}, such as triplet-loss~\cite{Hermans2017indensetriplet} and quadruple-loss~\cite{chen2017beyongdquaruplet}, aim to increase inter-identity feature distance and reduce intra-identity variation.
While many ReID methods excel in visible-visible tasks, 
they may struggle in low-light scenes due to inadequate handling of substantial domain gaps between visible and infrared modalities. Bridging this gap is crucial to enhance the versatility and applicability of ReID techniques across diverse environmental settings.

\subsection{Visible-Infrared Person Re-Identification}

The Visible-Infrared Person Re-identification (VI ReID) focuses on matching visible and infrared images of the same pedestrians across cameras. The query and gallery set are captured by cameras with different modalities. However, directly applying visible-visible ReID methods to VI ReID yielded poor results due to modality discrepancies and distinct distribution between modalities~\cite{ye2020bi}.

Modality compensation represents a VI ReID paradigm that initially employs a Generative Adversarial Network (GAN)~\cite{Goodfellow2014Generative} to generate another modality from the available one, then compensates the original image with the generated image, to mitigate the modality discrepancy. The cmGAN and D2RL~\cite{dai2018cmgancross,WANG2019D2RLLEARNING} are typical methodologies via modality compensation. 
Liu et al.~\cite{liu2022tsmerevisiting} proposed a Deep Jump Connection Generative Adversarial Network (DSGAN) to address the poor quality of generated images. 
In contrast to modality compensation methods~\cite{dai2018cmgancross,wang2019alignganrgb,WANG2019D2RLLEARNING,liu2022tsmerevisiting}, several methods leverage shared backbone to extract discriminative modality-invariant features from images of different modalities. The goal is to eliminate modality-specific information and thereby reduce the modality discrepancy. Chen et al.~\cite{CHEN2022SPOTstructure} proposed to model the modality invariant structural features of each modality, and use the information of human body structure and part position to learn discriminative cross-modality invariant features at the part level. Liang et al.~\cite{LIANG2023CMTRTMM} innovatively proposed to design modality embeddings for the ViT~\cite{dosovitskiy2020image} backbone and applied pure transformer~\cite{vaswani2017attention} networks to VI ReID task for the first time. 
Existing methods~\cite{CHEN2022SPOTstructure,wang2019alignganrgb} involve complex and computationally intensive feature alignment and fusion operations to address modality discrepancies. 
Some approaches~\cite{LU2023PMTlearning, li2021briding} normalize or optimize the training process to eliminate interference from modality-specific information on shared feature learning. 
For example, Lu et al.~\cite{LU2023PMTlearning} introduced grayscale images as auxiliary modalities and proposed a progressive learning strategy to help extract invariant features.
However, it is worth noting that modality-specific information may still contain discriminative details that are beneficial for VI ReID.

\subsection{Visual Prompt Learning}
Inspired by the success of textual prompts~\cite{brown2020language, lester2021thepower} in NLP, visual prompt learning has been widely used in several computer vision tasks. VPT~\cite{jia2022visual} designed a set of learnable vectors prepended to the input sequence of each encoder layer, thereby achieving better or equivalent results on 24 downstream recognition benchmarks. 
Bahng et al.~\cite{Bahng2022exploring} proposed to construct prompt in the form of perturbations, exploring the use of visual prompt guiding the large-scale language model CLIP~\cite{radford2021learning} to adapt to downstream tasks. Chen et al.~\cite{chen2022AdaptFormer} proposed to introduce learnable parameters into the pre-trained model for adapting to other downstream tasks, but different from the VPT, the learnable parameters are added to the transformer MLP layers rather than to the input sequence. Yang et al.~\cite{YANG2022prompting} and Zhu et al.~\cite{zhu2023visual} incorporated prompt learning into the multi-modality track field, showcasing its efficacy in multi-modality tasks. Yu et al.~\cite{yu2023prompting} and Liu et al.~\cite{liu2023promptbased} attempted to apply prompt in UDA ReID and DG ReID, respectively. These studies have demonstrated the potential of visual prompt learning in adapting original models to other tasks, but there is still no method to introduce visual prompt learning into the VI ReID field. In this work, we attempt to use specific prompts for different modality inputs and different instances to explore the application of visual prompt learning in the VI ReID.
\begin{figure*}[htbp] 
	\centering
	\includegraphics[width=0.95\linewidth,scale=0.95]{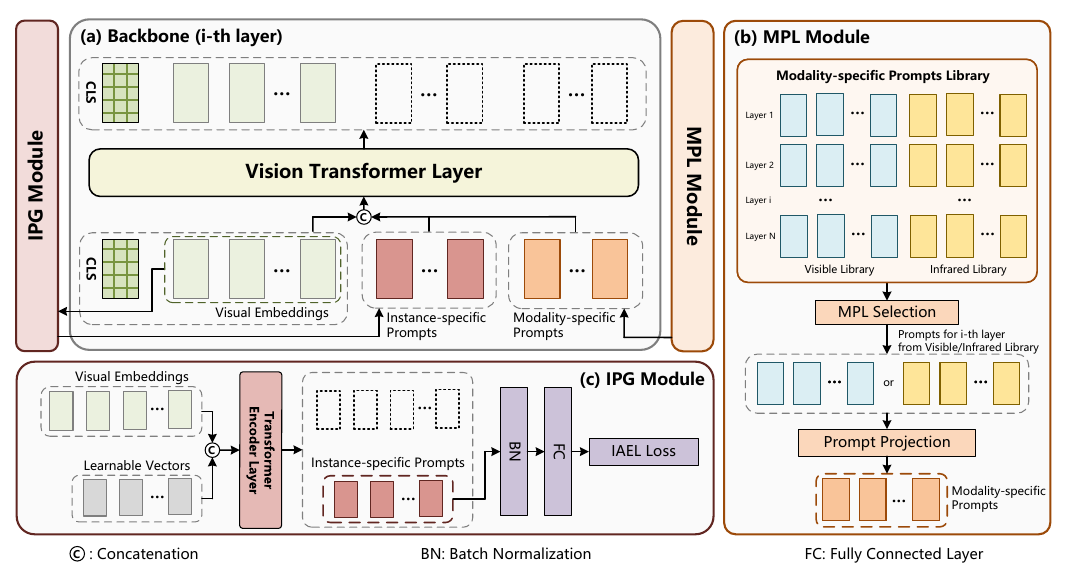}
        \vspace{-3mm}
	\caption{The overall framework of our proposed MIP network, which consists of a backbone model and two major modules. (a) A pre-trained vision transformer~\cite{dosovitskiy2020image} is used as the backbone model. (b) Modality-aware Prompts Learning (MPL) module produces modality-specific prompts for input visual embeddings of each layer according to the modality labels of input images. (c) Instance-aware Prompts Generator (IPG) module generates instance-specific prompts, and the generated prompts are supervised by "IAEL loss". The "IAEL Loss" is our proposed Instance-aware Enhancement Loss. The two kinds of rompts help the backbone network to adapt to different modality and instance inputs.
 }
	\label{fig2}
\end{figure*}

\section{Approach}
In this section, we give the details of our proposed Modality-aware and Instance-aware Visual Prompts (MIP) network. Firstly, we briefly introduce the overall framework of MIP. Thereafter, the major components of MIP, namely, the Modality-aware Prompt Learning (MPL) module and the Instance-aware Prompt Generator (IPG) module, are illustrated in sequence. Finally, we give the introduction of the objective functions employed in model training.

\subsection{Preliminary}
Our approach is built upon transformer architecture. In VI ReID, for an input image, we denote it as $x^m \in \mathbb{R}^{C\times H\times W}$, where the $C, H, W$ denote channel dimension, height, width of the images, and $m \in \{vis, ir\}$ denotes its modality flag. 
For each input image $x^m$, we split it into a patch sequence $s^m \in \mathbb{R}^{l\times C \times b\times b}$, where $l$ denotes the length of the sequence and $b$ denotes the size of the patch. 
Then, $s^m$ is converted to image embedding $e^m_0 \in \mathbb{R}^{l\times D}$ by a linear projection layer, where $D$ denotes the embedding dimension. 
Thereafter, an extra learnable class token $a_0$ will be concatenated with the embedding $e^m_0$, and $[a_0, e^m_0]$ will be sent into the transformer encoder layers as the initial input before the first layer, where $[\cdot]$ indicates the concatenation operation. 
Additionally, we denote the transformer encoder layers as $\{L_i\}^N_{i=1}$, where $N$ is the number of transformer encoder layers. Subsequently, the feature extraction stage of the $i$-th layer can be described as the following,
\[ [a_{i+1}, e^m_{i+1}] = L_i[a_{i}, e^m_{i}]\quad \tag{1} \]
where $a_{i+1}$ and $e^m_{i+1}$ denote the output class token and the output embedding of the $i$-th layer and $i = 1, ..., N$.

\subsection{Overall Framework}
In this paper, we introduce a MIP network designed for the VI ReID task. 
Our primary focus is on adapting the model to different modality inputs and instance inputs, thereby mining the correspondences between different modalities/instances to facilitate VI ReID. 
We achieve these goals by employing two distinct sets of visual prompts.
Specifically, we produce modality-specific prompts and instance-specific prompts according to current modality and instance input, and these two sets of prompts are concatenated after the feature embedding. 
By making use of modality-specific and instance-specific information preserved in these prompts, the model can explore potential correspondences between different modalities and instances, thereby facilitating the VI ReID. 

The overall framework of our MIP is depicted in Figure~\ref{fig2}. As shown in Figure~\ref{fig2}, we use a vision transformer pre-trained on ImageNet~\cite{deng2009imagenet} as the backbone and insert our designed modules, namely the Modality-aware Prompt Learning (MPL) module and the Instance-aware Prompt Generator (IPG) module, into the backbone model. These two proposed modules are responsible for capturing modality-specific and instance-specific information, and then producing modality-specific and instance-specific visual prompts, respectively. Then, the visual prompts will be concatenated with the input before each transformer encoder layer, like the following,
\[  [a_{i+1}, e^m_{i+1}, \_] = L_i[a_{i}, e^m_{i}, p_{i}] \\ \tag{2} \]
\[ p_{i} = [p^M_{i},p^I_{i}] \\  \tag{3} \]
where $p = [p^M_i, p^I_i]$ is the visual prompts, and the composition of modality-specific prompts $p^M_i$ and instance-specific $p^I_i$ will be specifically explained in sections 3.3 and 3.4; $L_i$ means the $i$-th transformer encoder layer and $i = 1, ..., N$; the $\_$ is the output of $p_{i}$, which is discarded.

\subsection{Modality-aware Prompt Learning}
The existing methods alleviate the interference of the modality gap by simply focusing on modality-invariant information while eliminating modality-specific information, overlooking the potential utilization of modality-specific information.
In this work, we notice that there could exist potential correspondences between modality-specific information, which could also assist re-identification. 
For an intuitive example, characteristics like clothing
types/materials, discerned from visible images, could influence the heat radiation intensities, thereby impacting infrared brightness variations.
Based on this idea, in this work, we attempt to guide the model to preserve and make use of modality-specific information.
In this step, we notice that many studies~\cite{YANG2022prompting, zhu2023visual, lu2022prompt, Bahng2022exploring, chen2022AdaptFormer}, e.g., VPT~\cite{jia2022visual} that achieved superior results across 24 downstream vision benchmarks, have demonstrated the potential of visual prompt learning in adapting original models to other tasks without complex operations and large extra computations, so we consider exploiting this advantage of prompt learning to help the model adapt to different modality inputs. We treat the adaptation of the model to visible and infrared modalities as two different tasks, applying two independent sets of learnable visual prompts for the two modalities, and also preserving the required modality-specific information. 
To achieve this, we propose an MPL module to produce modality-specific prompts, which is shown in Figure~\ref{fig2}(b).

Formally, the MPL module initializes and maintains a modality-specific prompts library with a set of learnable vectors. The modality-specific prompts library can be written as $P^M=\{p^{vis}_{i}, p^{ir}_{i}\}^N_{i=1}$, in which $N$ is the number of transformer encoder layers; $p_{i} \in \mathbb{R}^{{j} \times D}$;
the $j$ is the length of the modality-specific prompt, in this case, we set $j=16$; $p^{vis}_{i}$ and $p^{ir}_{i}$ denote the visible-specific and infrared-specific prompts for the input before the $i$-th layer. Before the input embedding is sent into each encoder layer, the MPL module selects the target modality-specific prompt from the library according to the modality label of the image and the index of the layer to be input. The chosen modality-specific prompt will be concatenated with the input after being projected by a linear layer. The above processing can be written as,
\[ \bar{p}^M_{i} = {\rm \Gamma}(P^M, m, i)  \tag{4}\]
\[ {\rm \Gamma}(P^M, m, i)= \begin{cases} p^{vis}_{i}\quad ,{\rm if}\ m=vis\\
          p^{ir}_{i}\quad ,{\rm if} \ m=ir \end{cases}  \tag{5}\]
\[ p^M_{i} = \varphi(\bar{p}^M_{i}) \tag{6}\]
where $i = 1, ..., N$; ${\rm \Gamma}(\cdot)$ denotes the MPL Selection operation; $\varphi(\cdot)$ denotes the prompt projection linear layer. 

\subsection{Instance-aware Prompt Generator}
In addition to tailoring the model for different modalities and exploring implicit correspondences among modality-specific information, we also strive to customize our model for distinct instances by leveraging instance-specific prompts.
The construction of instance-specific prompts is non-trivial.
Here, we notice a great challenge, namely, we cannot train a set of independent prompts for each instance, because the test instances are unlimited and unpredictable.
Therefore, in this work, we design a dynamic prompting module, capable of adaptively generating instance-specific prompts according to the input instance.
To accomplish this, shown in Figure~\ref{fig2}(c), we employ a transformer layer to transfer the instance-specific knowledge from the input image features into a group of learnable vectors.
The outputs of these learnable vectors then serve as instance-specific prompts to adapt our model to various input instances. 

Formally, given the input embedding $e^m_i$ before the $i$-th layer, we employ a transformer encoder layer to transfer the information in the embedding $e^m_i$ into a set of learnable vectors $v_i \in \mathbb{R}^{{k}\times D}$, where $i = 1,...,N$, and $k$ is the length of the instance-specific prompt, which is set as $k=16$ in this work. The process of constructing instance-specific prompt can be written as follows,
\[ [p^I_{i}, \_] = {\rm Trans}([v_i, e^m_{i}]) \tag{7} \]
where the $p^I_{i}$ is the output of $v_i$, i.e., instance-specific prompt; the ${\rm Trans}(\cdot)$ is a transformer encoder layer; the $\_$ is the output of $e^m_{i}$, which is discarded.

\subsection{Objective Function}
We employ our designed Instance-aware Enhancement Loss (IAEL) and some losses commonly used in ReID tasks for our model.

{\bfseries Instance-aware Enhancement Loss.} In practice, we observe that the modality-specific and instance-specific prompts are not easy to optimize. Particularly, the instance-specific prompts may degenerate to only get a trivial solution, which means it could be invariant to different instances while failing to be customized to them. To ensure that instance-specific prompts are indeed instance-customized, we have designed the IPG module in a generation-based manner rather than fusion-based ones like previous work~\cite{wang2022learning}, and we additionally designed a loss function called Instance Aware Enhancement Loss (IAEL) to force the generated prompts being instance-adaptive. The IAEL loss can be written formally as follows,
\[ L_{IAEL} = - \sum_{i=1}^N y\cdot \log\Bigl(\psi_i \bigl(\sigma_i (p^I_{i})\bigl)\Bigl) \tag{8} \]
where $i = 1,...,N$ means for the $i$-th layer; $y$ is the ground-truth label of current input instance; $\psi_i(\cdot)$ denotes the classifier for instance-specific prompts before the $i$-th layer; $\sigma_i(\cdot)$ denotes Batch Normalization~\cite{ioffe2015batchnorm} layer.

{\bfseries Overall Objective Function.} For model training, we adopt a hybrid loss function for our progressive learning framework. At the first stage, we utilize Cross-Entropy Loss $L_{ID}$ and Triplet Loss $L_{TRI}$,
\[ L_{1} =  L_{ID} + L_{TRI} \tag{9} \]

In the second stage, we further extract the reliable features with $L_{IAEL}$. The loss function can be defined as,
\[ L =  \alpha_1 L_1 + \alpha_2 L_{IAEL}\tag{10} \]
where the parameters $\alpha_1$ and $\alpha_2$ are hyperparameters, and we set $\alpha_1 = 1.0$ and $\alpha_2 = 0.5$. In practice, the model performs stably, and using different hyperparameters does not result in significant performance changes.
\section{Experiment}
In this section, we conduct comprehensive experiments to verify the superiority of our Modality-aware and Instance-aware Visual Prompts (MIP) network for VI ReID.
\begin{table*}
\renewcommand\arraystretch{1.05}
\large
\begin{center}
\caption{The experiment results of our MIP and other state-of-the-art methods under various test modes of SYSU-MM01 and RegDB datasets. Summarily, our proposed MIP outperforms other state-of-the-art methods on both two mainstream datasets. 
}
\label{tab1}
\vspace{-1mm}
\resizebox{1\textwidth}{!}{ 
    \begin{tabular}{c| c | c c | c c | c c | c c | c c | c c }
    \hline
    \multirow{4}*{Methods} & 
    \multirow{4}{*}{Reference} & 
    \multicolumn{8}{c|}{SYSU-MM01} & \multicolumn{4}{c}{\multirow{2}{*}{RegDB}} \\
    \cline{3-10}
    & & \multicolumn{4}{c|}{All-Search} & \multicolumn{4}{c|}{Indoor-Search} & \multicolumn{2}{c}{} & \multicolumn{2}{c}{} \\
    \cline{3-14}
    & & \multicolumn{2}{c|}{Single-Shot} & \multicolumn{2}{c|}{Multi-Shot} & \multicolumn{2}{c|}{Single-Shot} & \multicolumn{2}{c|}{Multi-Shot} & \multicolumn{2}{c|}{Infrared to Visible} & \multicolumn{2}{c}{Visible to Infrared} \\
    \cline{3-14}
    & & Rank-1 & mAP & Rank-1 & mAP & Rank-1 & mAP & Rank-1 & mAP & Rank-1 & mAP & Rank-1 & mAP \\
    \hline \hline
    Zero-Pad~\cite{wu2017rgb} & ICCV'17 & 14.80 & 15.95 & 19.13 & 10.89 & 20.58 & 26.92 & 24.43 & 18.64 & - & - & - & - \\
    cmGAN~\cite{dai2018cmgancross} & IJCAI'18 & 26.97 & 27.80 & 31.49 &  22.27 & 31.63 & 42.19 & 37.00 & 32.76 & - & - & - & - \\
    AlignGAN~\cite{wang2019alignganrgb} & ICCV'19 & 42.40 & 40.70 & 51.50 & 33.90 & 45.90 & 54.30 & 57.10 & 45.30 & 56.30 & 53.40 & 57.90 & 53.60 \\
    \hline
    CMM+CML~\cite{ling2020cmmcmlclass} & MM'20 & 51.80 & 51.21 & 56.27 & 43.39 & 54.98 & 63.70 & 60.42 & 53.52 & 59.81 & 60.86 & - & - \\
    DDAG~\cite{YE2020DDAGdynamic} & ECCV'20 & 54.75 & 53.02 & - & - & 61.02 & 67.98 & - & - & 68.06 & 61.80 & 69.34 & 63.46 \\
    \hline
    NFS~\cite{chen2021nfsneural} & CVPR'21 & 56.91 & 55.45 & 63.51 & 48.56 & 62.79 & 69.79 & 70.03 & 61.45 & 77.95 & 69.79 & 80.54 & 72.10 \\
    CM-NAS~\cite{FU2021CMNAS} & ICCV'21 & 61.99 & 60.02 & 68.68 & 53.45 & - & - & - & - & 84.54 & 80.32 & 82.57 & 78.31\\
    AGW~\cite{YEMANG2021AGWdeep} & TPAMI'21 & 47.50 & 47.65 & - & - & 54.17 & 62.97 & - & - & - & - & 70.05 & 66.37 \\
    \hline
    MID~\cite{FU2022MIDmodality} & AAAI'22 & 60.27 & 59.40 & - & - & 64.86 & 70.12 & - & - & 84.29 & 81.41 & 87.45 & \underline{84.85} \\
    FMCNet~\cite{ZHANG2022FMCNET} & CVPR'22 & 66.34 & 62.51 & - & - & 68.15 & 74.09 & - & - & 88.38 & \underline{83.86} & 89.12 & 84.43\\
    SPOT~\cite{CHEN2022SPOTstructure} & TIP'22 & 65.34 & 62.25 & - & - & 69.42 & 74.63 & - & - & 79.37 & 72.26 & 80.35 & 72.46\\
    \hline
    PMT~\cite{LU2023PMTlearning} & AAAI'23 & \underline{67.53} & \underline{64.98} & - & - & 71.66 & 76.52 & - & - & 84.16 & 75.13 & 84.83 & 76.55\\
    TOPLight ~\cite{YU2023TOPLIGHT} & CVPR'23 & 66.76 & 64.01 & - & - & \underline{72.89} & \underline{76.70} & - & - & 80.65 & 75.91 & 85.51 & 79.95\\
    CMTR~\cite{LIANG2023CMTRTMM} & TMM'23 & 65.45 & 62.90 & \underline{71.99} & \underline{57.07} & 71.46 & 76.67 & \underline{80.00} & \underline{69.49} & 84.92 & 80.79 & 88.11 &  81.66\\
    DFLN-ViT ~\cite{ZHAO2023SPATIAL-DFLN-VIT} & TMM'23 & 59.54 & 57.70 & - & -& 62.13 & 69.03 & - & - & \underline{91.21} & 81.62 & \textbf{92.10} & 82.11\\
    \hline
    \textbf{OURS} & - & \textbf{70.84} & \textbf{66.41} & \textbf{74.68} & \textbf{58.49} & \textbf{78.80} & \textbf{79.92} & \textbf{82.29} & \textbf{70.65} & \textbf{92.38} & \textbf{85.99} & \underline{91.26} & \textbf{85.90} \\
    \hline
    \end{tabular}
    }
\end{center}
\end{table*}

\subsection{Datasets and Evaluation Protocols}
Two mainstream public VI ReID datasets, SYSU-MM01 and RedDB, are used for our experiments.

{\bfseries SYSU-MM01}~\cite{wu2017rgb} consists of 491 persons whose 286,628 visible and 15,792 infrared images are captured by 4 visible and 2 infrared cameras. The dataset is divided into the training set and the test set. There are 22,258 visible images and 11,909 infrared images of 395 persons in the training set, and the test set contains images of 96 persons whose 3,803 infrared images are used as query images and several visible images are randomly selected as gallery images. According to the general setting, the composition of gallery images is decided by two search modes: {\slshape All-Search} and {\slshape Indoor-Search}, and two selection modes: {\slshape Single-Shot} and {\slshape Multi-Shot}. 
In {\slshape All-Search} mode, images under all visible cameras can be selected, while in {\slshape Indoor-Search} mode, only images under indoor visible cameras can be selected for gallery images. And 1 image is selected for each person in {\slshape Single-Shot} mode, and 10 images are selected for each person in {\slshape Multi-Shot} mode.

{\bfseries RegDB}~\cite{nguyen2017person} consists of 412 persons whose 4,120 visible and 4,120 infrared images are captured by one visible camera and one infrared camera, and there are 10 visible and 10 infrared images for each person. The training set includes the images of 206 randomly chosen persons and the test set contains the images of the remaining 206 persons.
In the test mode {\slshape Visible to Infrared}, the visible images are used as query images, and the infrared images are used as gallery images, and the arrangement is reversed in {\slshape Infrared to Visible} mode.

{\bfseries Evaluation Protocols.} We adopt the Cumulative Matching Characteristic curve (CMC) and mean Average Precision (mAP) as the evaluation metrics on the two datasets, to quantitatively evaluate the performance of our proposed model. For CMC, we calculate the percentage of correctly retrieved images among top-1 results (Rank-1 accuracy) based on similarity.

\subsection{Implementation Details}
Our proposed MIP method is implemented on the Pytorch framework, and one NVIDIA RTX3090 GPU is used to execute all experiments. In our experiments, we use a vision transformer~\cite{dosovitskiy2020image} pre-trained on ImageNet~\cite{deng2009imagenet} as the backbone. During training, the input images are resized to $256\times 128$, and the data augmentations, including random cropping, color jittering, random erasing, and grayscale, are applied to transform the input images. The batch size is set to 4 (persons) $\times$ (8 (visible images) + 8 (infrared images)) = 64. We adopt the Stochastic Gradient Descent (SGD) optimizer to train the model for 80 epochs with an initial learning rate of $1.0 \times 10^{-2}$. The learning rate will decay with the cosine decay schedule, where the minimum learning rate is $1.0 \times 10^{-5}$, the momentum parameter is 0.9, and the decay rate is 0.1. 
Following the framework of Figure 2, we maintain two sets of modality-specific prompts for each layer and a shared IPG module for all layers. This results in 7.67\% and 4.09\% extra parameters, respectively.

\subsection{Comparison with State-of-the-art methods}
We compare our MIP with existing state-of-the-art methods for VI ReID, in mainstream datasets SYSU-MM01 and RegDB. The comparison results are reported in Table~\ref{tab1}, where bold and underlined fonts indicate the best and second-best performance, respectively.

{\bfseries SYSU-MM01:} We can observe that our proposed MIP method shows the best results on all the metrics under all SYSU-MM01 dataset modes.
Compared with other transformer-based methods, PMT, CMTR, DFLN-ViT, SPOT~\cite{CHEN2022SPOTstructure} and recent CNN-based method TOPLight~\cite{YU2023TOPLIGHT}, our MIP outperforms the state-of-the-art results by 3.31\%/5.91\% Rank-1 and 1.43\%/3.22\% mAP under the {\slshape All-Search}/ {\slshape Indoor-Search} + {\slshape Single-Shot} mode. Recent other methods are lacking rich experiments under changeable {\slshape Multi-Shot} mode, and MIP still achieves better results than CMTR with an improvement of 2.69\%/2.29\% Rank-1 and 1.42\%/1.16\% mAP.

{\bfseries RegDB:} As shown in Table~\ref{tab1}, our proposed MIP method also achieves promising performance on RegDB.
Compared with other top-performing methods, MIP outperforms the state-of-the-art results by 1.17\% Rank-1 and 2.13\% mAP under {\slshape Infrared to Visible} mode. MIP also outperforms the state-of-the-art result by 1.05\% mAP under {\slshape Visible to Infrared} mode, but DFLN-ViT shows the best Rank-1 result of 92.10\%, while MIP also shows a comparable Rank-1 result of 91.26\% and outperforms by 3.79\% mAP. The main reason for DFLN-ViT's better result could be that its structure-aware mechanism can model align persons' structures easily on simple images in RegDB. But this kind of method is viewpoint sensitive, and its performance on SYSU-MM01 with complex viewpoint changes is 11.89\% lower than MIP on average.


\subsection{Ablation Studies}
In this subsection, we conduct a series of ablation experiments to evaluate the effectiveness of our proposed MIP. 
We begin by validating the effectiveness of each component. Next, we compare our carefully designed MPL and IPG modules with general prompt-based approaches. Subsequently, we discuss the necessity of the generation-based design of the IPG module. Finally, we present the effect of the MPL module on extracting implicit correspondence between modality-specific information.
Notably, all the ablation experiments are conducted on the SYSU-MM01 dataset under {\slshape Single-Shot} mode, except when specifically stated. 

{\bfseries Effectiveness of Proposed Components.} 
To evaluate the effectiveness of our proposed MPL module, IPG module, and IAEL loss, we add these three components to the baseline gradually and evaluate the performances. 
As shown in Table~\ref{tab2}, compared with Method-1, i.e., baseline model, Method-2 is trained with an additional MPL module, which brings a significant improvement of 22.44\% mAP and 27.40\% Rank-1 on average under two search modes on SYSU-MM01 dataset.
Meanwhile, Method-3 adopts baseline+IPG as the model and improves the performance of the baseline model by 14.23\% mAP and 16.44\% Rank-1 on average on SYSU-MM01.
Based on Method-3, Method-4 adds IAEL loss to promote the prompt generated by the IPG module to be more customized to each input instance, achieving an average improvement of 2.59\% mAP and 3.44\% Rank-1 compared with Method-3.
What’s more, Method-5 equips the baseline with both MPL and IPG modules, which brings a further improvement of 8.30\% mAP and 9.04\% Rank-1 on average than baseline+MPL and baseline+IPG. This indicates that our designed modules can complement to jointly promote performance improvement.
Finally, Method-6, our full model (MIP), achieves an additional average improvement of 1.51\% in mAP and 2.74\% in Rank-1 compared to Method-5, thanks to the enhancement from the IAEL loss.
We also conduct the same ablation experiments on the RegDB dataset, as shown in Table~\ref{tab3}, and the addition of individual components similarly brings performance improvements to the model.
From these experiments, we can find that the MPL module, IPG modules, and IAEL Loss are effective in adapting the model to different modalities and instances and enhancing its capability for VI ReID.

\begin{table}
\renewcommand\arraystretch{1.03}
\begin{center}
\caption{The effects of our proposed components performed on SYSU-MM01. 
}
\label{tab2}
\vspace{-3mm}
\resizebox{0.485\textwidth}{!}{ 
    \begin{tabular}{c| c c c | c c | c c }
    \hline
    \multirow{2}*{Methods} & 
    \multicolumn{3}{c|}{Components} & \multicolumn{2}{c|}{All-Search} & \multicolumn{2}{c}{Indoor-Search} \\
    \cline{2-8}
    & MPL & IPG & IAEL & Rank-1 & mAP & Rank-1 & mAP  \\
    \hline \hline
    1(Base)     &           &           &           & 40.49 & 40.93 & 41.76 & 49.12\\
    \hline
    2           & \ding{52} &           &           & 63.77 & 60.10 & 73.28 & 74.83\\
    3           &           & \ding{52} &           & 52.98 & 52.75 & 62.14 & 65.75\\
    4           &           & \ding{52} & \ding{52} & 55.48 & 54.81 & 66.53 & 68.87\\
    5           & \ding{52} & \ding{52} &           & 67.21 & 64.35 & 76.95 & 78.95\\
    \hline
    \textbf{6(Full)}     & \ding{52} & \ding{52} & \ding{52} & \textbf{70.84} & \textbf{66.41} & \textbf{78.80} & \textbf{79.92}\\
    \hline
    \end{tabular}
    }
\end{center}
\end{table}

\begin{table}
\renewcommand\arraystretch{1.06}
\begin{center}
\caption{The effects of our proposed components performed on RegDB.
}
\label{tab3}
\vspace{-3mm}
\resizebox{0.485\textwidth}{!}{ 
    \begin{tabular}{c| c c c | c c | c c }
    \hline
    \multirow{2}*{Methods} & 
    \multicolumn{3}{c|}{Components} & \multicolumn{2}{c|}{Infrared to Visible} & \multicolumn{2}{c}{Visible to Infrared} \\
    \cline{2-8}
    & MPL & IPG & IAEL & Rank-1 & mAP & Rank-1 & mAP  \\
    \hline \hline
    1(Base)     &           &           &           & 87.14 & 80.20 & 85.05 & 79.16\\
    \hline
    2           & \ding{52} &           &           & 88.64 & 82.89 & 89.81 & 82.27\\
    3           &           & \ding{52} &           & 88.70 & 83.20 & 89.50 & 82.80\\
    4           &           & \ding{52} & \ding{52} & 89.85 & 84.05 & 89.90 & 83.47\\
    5           & \ding{52} & \ding{52} &           & 91.26 & 85.43 & 91.17 & 84.86\\
    \hline
    \textbf{6(Full)}     & \ding{52} & \ding{52} & \ding{52} & \textbf{91.26} & \textbf{85.90} & \textbf{92.38} & \textbf{85.99}\\
    \hline
    \end{tabular}
    }
\end{center}
\end{table}

\begin{table}
\renewcommand\arraystretch{1.03}
\begin{center}
\caption{The effects of our designed MPL and IPG modules compared with general prompt-based approaches. The "\ding{52}\ding{52}" means using two sets of general prompts to replace modality-specific and instance-specific prompts, respectively.}
\label{tab4}
\vspace{-3mm}
\resizebox{0.485\textwidth}{!}{ 
    \begin{tabular}{c| c c c | c c | c c }
    \hline
    \multirow{2}*{Methods} & 
    \multicolumn{3}{c|}{Prompts} & \multicolumn{2}{c|}{All-Search} & \multicolumn{2}{c}{Indoor-Search} \\
    \cline{2-8}
    & General & MPL & IPG &  Rank-1 & mAP & Rank-1 & mAP  \\
    \hline \hline
    1(Base)     &&           &           &            40.49 & 40.93 & 41.76 & 49.12\\
    \hline
    2           & \ding{52} &           &                      & 44.73 & 42.73 & 54.62 & 58.48\\
    3           &           & \ding{52} &                      & 63.77 & 60.10 & 73.28 & 74.83\\
    4           &           &           & \ding{52}            & 52.98 & 52.75 & 62.14 & 65.75\\
    \hline
    5           & \ding{52}\ding{52} &           &                      & 51.70 & 47.44 & 57.29 & 60.17\\
    6           &           & \ding{52} & \ding{52}            & 67.21 & 64.35 & 76.95 & 78.95\\
    \hline
    \hline
    \end{tabular}
    }
\end{center}
\end{table}

{\bfseries Comparisons with General Prompt-based Approaches.} 
Several prompt-based approaches~\cite{jia2022visual, YANG2022prompting, zhu2023visual, lu2022prompt, Bahng2022exploring, chen2022AdaptFormer} have demonstrated the ability of visual prompt learning in adapting original models to various tasks. 
To explore whether the advantages of MPL and IPG modules solely depend on effective visual prompts learning, the modules designed in MIP are replaced with general prompts learning~\cite{jia2022visual}, and the experiments are repeated, results of which are shown in Table~\ref{tab4}. 

As shown in Table~\ref{tab4}, Method-2, i.e., baseline+general-prompts, only improves performance with an average increase of 5.58\% mAP and 8.55\% Rank-1 compared to the baseline model. 
In contrast, in Method-3 and Method-4, we replace the general prompts learning with our designed MPL module and IPG module, introducing significant improvements to Method-2, i.e., 16.86/8.65\% mAP and 18.85/7.89\% Rank-1 on average, respectively. This result illustrates the necessity to design specifically for modality and instance adaptation. 
Furthermore, the performance of Method-6 equipped with both MPL and IPG is 17.85\%  higher in mAP and 17.59\% higher in Rank-1 than Method-5 using two sets of general prompts, which achieves more obvious advantages than that when only equipped with a single module.

The above ablation experiments could denote that the improvement brought by MPL and IPG modules for VI ReID tasks is not solely dependent on the effectiveness of visual prompt learning. We illustrate the necessity of carefully designing these two modules to help the model adapt to different modalities and different instances and the necessity of capturing modality-specific and instance-specific information to mine potential relationships.

\begin{table}[t] 
\renewcommand\arraystretch{1.06}
\begin{center}
\caption{The comparisons of generation-based and fusion-based IPG modules. 
}
\label{tab5}
\vspace{-3mm}
\resizebox{0.482\textwidth}{!}{ 
    \begin{tabular}{c| c c | c | c | c c | c c }
    \hline
    \multirow{2}*{Methods} & 
    \multicolumn{2}{c|}{IPG} & \multirow{2}*{IAEL} & \multirow{2}*{MPL} & \multicolumn{2}{c|}{All-Search} & \multicolumn{2}{c}{Indoor-Search} \\
    \cline{2-3}\cline{6-9}
    & Fusion & Generation & & & Rank-1 & mAP & Rank-1 & mAP  \\
    \hline \hline
    1(Base)     &           &           &           & \multirow{5}*{} & 40.49 & 40.93 & 41.76 & 49.12\\
    \cline{1-3}\cline{6-9}
    2           & \ding{52} &           &           && 54.17 & 50.01 & 62.73 & 64.42\\
    3           &           & \ding{52} &           && 52.98 & 52.75 & 62.14 & 65.75\\
    \cline{1-4}\cline{6-9}
    4           & \ding{52} &           & \multirow{2}*{\ding{52}} && 56.30 & 52.87 & 67.98 & 69.27\\
    5           &           & \ding{52} &                          && 55.48 & 54.81 & 66.53 & 68.87\\
    \hline
    6           &           &           &           & \multirow{5}*{\ding{52}} & 63.77 & 60.10 & 73.28 & 74.83\\
    \cline{1-3}\cline{6-9}
    7           & \ding{52} &           &           && 63.00 & 59.86 & 74.05 & 74.98\\
    8           &           & \ding{52} &           && 67.21 & 64.35 & 76.95 & 78.95\\
    \cline{1-4}\cline{6-9}
    9           & \ding{52} &           & \multirow{2}*{\ding{52}} && 63.11 & 59.90 & 73.19 & 74.49\\
    \textbf{10(Full)}     &           & \ding{52} &&& \textbf{70.84} & \textbf{66.41} & \textbf{78.80} & \textbf{79.92}\\
    \hline
    \end{tabular}
    }
\end{center}
\end{table}

\begin{figure}[ht]
	\centering
	\includegraphics[width=0.9\linewidth,scale=0.95]{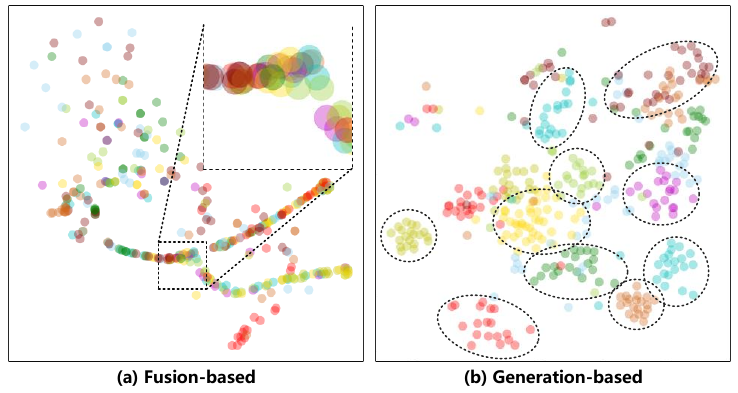}
        \vspace{-2mm}
	\caption{The t-SNE visualizations results of prompts from fusion-based and generation-based IPG modules. Different colors represent distinct identities. (a) Fusion-based IPG prompts cluster closely, with less obvious boundaries between individuals, indicating weaker instance-aware ability. (b) Generation-based IPG prompts show increased distances between individuals, reflecting stronger instance-aware ability, crucial for effective adaptation to diverse instances. }

	\label{fig3}
\end{figure}

\begin{figure}[t] 
	\centering
	\includegraphics[width=0.9\linewidth,scale=0.95]{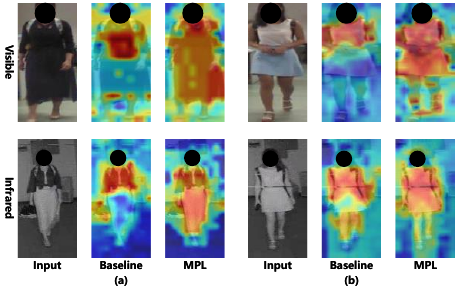}
        \vspace{-2mm}
	\caption{The visualizations results of attention maps of our MPL module and baseline model. From the second column in each case, we could find that the baseline model tends only to capture the explicit correspondence between different modality inputs. Such as only focusing on the upper dress part while ignoring the skirt part in case (a). As for our MPL module, with the carefully designed modality-specific prompts, it could effectively adapt to and make use of the modality-specific information. This enables our MPL model to explore and capture the implicit correspondence between the skirts part. Case (b) shows a similar result. }
 

	\label{fig4}
\end{figure}

{\bfseries Ablation about the product strategy of instance-specific prompts.} 
Unlike our generation-based strategy, previous work~\cite{wang2022learning} proposes a fusion-based instance-specific prompt learning pipeline. This pipeline generates specific weights for each input instance, which are then used to fuse several prompt prototypes and create instance-specific prompts. 
To validate the necessity of our generation-based strategy, we implement a fusion-based IPG module and conduct several experiments to compare it with our generation-based IPG module.

As shown in Table~\ref{tab5}, Method-7, which adds a fusion-based IPG module to Method-6, does not show additional improvement over Method-6.
In contrast, Method-8, which adds a generation-based IPG module to Method-6, demonstrates a significant improvement.
After introducing the IAEL loss, comparing Method-6, Method-9, and Method-10 in Table~\ref{tab5}, we can also get a similar finding.
In addition, Method-1 to Method-5 show the results without the MPL module. We can find that the fusion-based IPG module brings an average improvement of 12.19\% mAP and 17.33\% Rank-1 to the baseline when it does not work with the MPL module. However, its improvement of mAP is still 2.04\% lower than the improvement brought by the generation-based IPG module on average. The results are similar after introducing the IAEL loss.
The reason behind this could be that, the awareness ability of prompts generated by the fusion-based IPG module may not be strong enough because of degenerating to the trivial solution. Consequently, when the two modules work together, the fusion-based IPG module might duplicate the role of the MPL module.

To intuitively explore this finding, we exhibit the t-SNE visualization results of instance-specific prompts produced by fusion-based and generation-based IPG modules in Figure~\ref{fig3}. We randomly sample 386 images of 10 different persons. The different colors of dots mean different persons. As shown in Figure~\ref{fig3}, the prompts produced by generation-based IPG have obvious intra-person distances and clear boundaries, while the prompts generated by the fusion-based IPG are distributed in a disordered and interleaving manner, which indicates that prompts produced by generation-based IPG have more fine-grained instance-aware ability than prompts produced by fusion-based IPG.

{\bfseries Focused Contents of the MPL Module.} 
We present some visualization cases to further analyze what the MPL module focuses on and determine if MPL does extract the implicit correspondence between modality-specific information. As shown in Figure~\ref{fig4}, we generate the attention maps of some case images in baseline and model with the MPL module. 
Taking case (a) in Figure~\ref{fig4} as an example, as shown in the first column, the color and brightness of skirts presented in different modalities are significantly different.
Therefore, from the second column, we find that the baseline model tends to only capture the explicit correspondence between different modality inputs, thereby focusing only on the upper dress part while ignoring the skirt part. 
As for our MPL module, with the carefully designed modality-specific prompts, it could effectively adapt to and make use of the modality-specific information.
This enables our MPL model to explore and capture the implicit correspondence between the skirts part.
This is consistent with our motivation, and case (b) shows a similar result.

\section{Conclusion}
In this work, we propose a novel VI ReID method, the Modality-aware and Instance-aware Visual Prompts (MIP) network, which incorporates visual prompt learning into the VI ReID field.
We design a Modality-aware Prompt Learning (MPL) module and an Instance-aware Prompt Generator (IPG) to generate modality-specific and instance-specific prompts for the ReID model, which guides the model to adapt to inputs of different modalities and leverage modality-specific information, and capture identity-specific discriminative clues that help re-identification, respectively.
We anticipate that this work will serve as a catalyst for future research exploring the significance of modality-specific and instance-specific information in VI ReID.

\begin{acks}
This work was supported by the National Natural Science Foundation of China (No.U23B2013), the Shaanxi Provincial Key R\&D Program (No.2021KWZ-03, No.2024GX-YBXM-117), and the Natural Science Basic Research Program of Shaanxi (No.2021JCW-03).
\end{acks}

\clearpage

\bibliographystyle{ACM-Reference-Format}
\balance
\bibliography{sample-base}

\end{document}